\documentclass[review]{elsarticle}

\usepackage{lineno,hyperref}
\usepackage{subcaption}
\usepackage{amsmath,amssymb}
\usepackage{mathtools}
\usepackage{amsthm}

\usepackage[export]{adjustbox}
\usepackage[ruled,vlined]{algorithm2e}
\usepackage{graphbox}
\usepackage{color,soul}
\usepackage{lipsum}
\usepackage{makecell}
\usepackage{graphics}
\usepackage{color,soul}
\usepackage{xcolor, colortbl} 

\usepackage{amsmath,amsfonts,bm}

\newcommand{\RNum}[1]{\uppercase\expandafter{\romannumeral #1\relax}}
\newcommand{\Rnum}[1]{\lowercase\expandafter{\romannumeral #1\relax}}

\def\eqref#1{equation~(\ref{#1})}
\def\Eqref#1{Equation~(\ref{#1})}

\def\0{\bm{0}} 
\def\1{\bm{1}}

\def\vdelta{{\bm{\delta}}} 

\def\vn{{\bm{n}}}

\def\vx{{\bm{x}}}
\def\vy{{\bm{y}}}

\DeclareMathAlphabet{\mathsfit}{\encodingdefault}{\sfdefault}{m}{sl}
\SetMathAlphabet{\mathsfit}{bold}{\encodingdefault}{\sfdefault}{bx}{n}

\usepackage{diagbox}
\usepackage{multirow}
\usepackage{subcaption}
\usepackage[export]{adjustbox}
\usepackage{amssymb}

\modulolinenumbers[100]
\hypersetup{colorlinks = true, linkcolor = blue, anchorcolor =red, citecolor = blue, filecolor = red, urlcolor = red, pdfauthor=author}
\journal{Journal of \LaTeX\ Templates}

\usepackage{framed}
\colorlet{shadecolor}{yellow!40}

\let\oldequation\equation
\let\oldendequation\endequation

\renewenvironment{equation}
  {\linenomathNonumbers\oldequation}
  {\oldendequation\endlinenomath}
  
\begin{document}

\begin{frontmatter}

\title{Comment on Transferability and Input Transformation with Additive Noise} 


\author[mymainaddress]{Hoki Kim}

\author[mymainaddress]{Jinseong Park}

\author[mymainaddress]{Jaewook Lee\corref{mycorrespondingauthor}}
\cortext[mycorrespondingauthor]{Corresponding author}
\ead{jaewook@snu.ac.kr}

\address[mymainaddress]{Seoul National University, Gwanakro 1, Seoul, South Korea}

\begin{abstract}
Adversarial attacks have verified the existence of the vulnerability of neural networks. By adding small perturbations to a benign example, adversarial attacks successfully generate adversarial examples that lead misclassification of deep learning models. More importantly, an adversarial example generated from a specific model can also deceive other models without modification. We call this phenomenon ``transferability". Here, we analyze the relationship between transferability and input transformation with additive noise by mathematically proving that the modified optimization can produce more transferable adversarial examples.
\end{abstract}


\end{frontmatter}

\linenumbers

Given an input $\vx$, a one-hot encoded vector $\vy$ and classifier $h$, adversarial attack \cite{szegedy2013intriguing, goodfellow2014explaining} maximizes a loss function $\mathcal{L}$ with an adversarial perturbation $\vdelta$ as follows:
\begin{equation}
    \max_{\|\vdelta\| \leq \epsilon} \mathcal{L}(h(\vx+\vdelta), \vy)
\label{eq:minmax}
\end{equation}
where $\epsilon$ is the maximum size of $\delta$. To achieve high transferability, we should maximize the expectation of the loss over the hypothesis space $\mathcal{H}$.
\begin{equation}
\max_{\|\vdelta\| \leq \epsilon}f(\vx+\vdelta):=\mathbb{E}_{h\sim \mathcal{H}}\big[\mathcal{L}(h(\vx+\vdelta), \vy)\big]
\label{eq:hightrans}
\end{equation}
Since the hypothesis space $\mathcal{H}$ is generally unknown, existing methods try to find the adversarial example $\vx'= \vx + \vdelta$ based on the sampled model (or source model) $h$ as follows:
\begin{equation}
\max_{\|\vdelta\| \leq \epsilon}\hat{f}(\vx+\vdelta):=\mathcal{L}(h(\vx+\vdelta), \vy)
\label{eq:samplef}
\end{equation}
However, as described in the main text, \Eqref{eq:samplef} has the generalization problem as it generates overfitted adversarial examples to the source model $h$.

Now, we consider the following maximization optimization:
\begin{equation}\label{eq:proposed}
\max_{\|\vdelta\| \leq \epsilon}\mathbb{E}_{\vn \sim \mathcal{N}}\big[\mathcal{L}(h(\vx+\vdelta+\vn), \vy)\big]
\end{equation}
To optimize \Eqref{eq:proposed}, given $M$ sampled noises $\vn_m\sim \mathcal{N}$ where $m=1,...,M$, we instead maximize the following equation:
\begin{equation}
\max_{\|\vdelta\| \leq \epsilon} \bar{f}(\vx+\vdelta):= \sum_{m=1}^M\alpha_m f_m(\vx+\vdelta) \label{eq:USE}
\end{equation}
where $f_m(\vx+\vdelta):=\mathcal{L}(h(\vx+\vdelta+\vn_m), \vy)$, $\alpha_m\geq 0$ and $\sum_{m=1}^M\alpha_m=1$.

Now, we mathematically prove that the objective of the proposed method $\bar{f}$ can provide better alignment with the objective $f$ than that of existing methods $\hat{f}$ by establishing the expectation of the mean squared error. The error of the loss can be formalized as follows:
\begin{equation*}
\hat{\varepsilon}(\vx+\vdelta) = \hat{f}(\vx+\vdelta) - f(\vx+\vdelta),\quad
\varepsilon_m(\vx+\vdelta) = f_m(\vx+\vdelta) - f(\vx+\vdelta).
\end{equation*}
Here, we assume that, with a proper noise distribution $\mathcal{N}$, the hypothesis space $\mathcal{H}$ is large enough so that $h(\cdot+\vn)\in \mathcal{H}$ for any $\vn\sim\mathcal{N}$ whenever $h(\cdot)\in \mathcal{H}$. This assumption can be supported by the fact that vanilla trained models have decent performance for noised examples corrupted by small Uniform or Gaussian noise and the existence of noise-robust training such as adversarially trained models under the allowable maximum perturbation $\epsilon$ \cite{madry2017towards, wong2020fast}. From these experimental results, we can expect that the hypothesis space $\mathcal{H}$ is large enough to include $h(\cdot+\vn)$ for $h(\cdot)\in \mathcal{H}$ in the general training framework.


Then, by the assumption, $\mathbb{E}_{h\sim\mathcal{H}}\big[g(h(\vx+\vdelta+\vn_m), \vy)\big]=\mathbb{E}_{h_m\sim\mathcal{H}}\big[g(h_m(\vx+\vdelta), \vy)\big]$ for any function $g(\cdot)$. Thus,  $\mathbb{E}_{h\sim\mathcal{H}}\big[\varepsilon_m(\vx+\vdelta)^2\big]=\mathbb{E}_{h\sim\mathcal{H}}\big[\hat{\varepsilon}(\vx+\vdelta)^2\big]$.
Now, we calculate the mean objective squared error between $\bar{f}(\vx+\vdelta)$ and $f(\vx+\vdelta)$.
\begin{eqnarray*}
\mathbb{E}_{h\sim\mathcal{H}}\large\big[(\bar{f}(\vx+\vdelta)-f(\vx+\vdelta))^2\large\big]=
\mathbb{E}_{h\sim\mathcal{H}}\large\big[(\sum_{m=1}^M\alpha_m \varepsilon_m(\vx+\vdelta))^2\large\big]\\ =\mathbb{E}_{h\sim\mathcal{H}}\large\big[\sum_{m=1}^M\alpha_m^2 \varepsilon_m(\vx+\vdelta)^2 +2\sum_{l=1}^M\sum_{k=l+1}^M \varepsilon_l(\vx+\vdelta)\varepsilon_k(\vx+\vdelta)\alpha_l \alpha_k\large\big].
\end{eqnarray*}
By the Cauchy-Schwartz inequality,
\begin{eqnarray*}
\mathbb{E}_{h\sim\mathcal{H}}\big[\varepsilon_l(\vx+\vdelta)\varepsilon_k(\vx+\vdelta)\big]
\leq  \mathbb{E}_{h\sim\mathcal{H}}\big[\varepsilon_l(\vx+\vdelta)^2\big]^{\frac12}
\mathbb{E}_{h\sim\mathcal{H}}\big[\varepsilon_k(\vx+\vdelta)^2\big]^{\frac12}=\mathbb{E}_{h\sim\mathcal{H}}\big[\hat{\varepsilon}(\vx+\vdelta)^2\big]
\end{eqnarray*}
for $l,k=1,...,M$. Thus, we can define $\rho_{l,k}$ with $0\leq \rho_{l,k} \leq 1$ and $\rho_{l,l}=1$ by
\begin{eqnarray*}
 \mathbb{E}_{h\sim\mathcal{H}}\big[\varepsilon_l(\vx+\vdelta)\varepsilon_k(\vx+\vdelta)\big]= \rho_{l,k} \mathbb{E}_{h\sim\mathcal{H}}\big[\hat{\varepsilon}(\vx+\vdelta)^2\big].
\end{eqnarray*}
Therefore, the following inequality holds.
\begin{eqnarray*}
\mathbb{E}_{h\sim\mathcal{H}}\big[(\bar{f}(\vx+\vdelta)-f(\vx+\vdelta))^2\big]=\big(\sum_{m=1}^M\alpha_m^2  +2\sum_{l=1}^M\sum_{k=l+1}^M \rho_{l,k}\alpha_l\alpha_k\big)\cdot \mathbb{E}_{h\sim\mathcal{H}}\big[\hat{\varepsilon}(\vx+\vdelta)^2\big]\\
\leq  \big(\sum_{m=1}^M\alpha_m^2  +2\sum_{l=1}^M\sum_{k=l+1}^M \alpha_l\alpha_k\big)\cdot \mathbb{E}_{h\sim\mathcal{H}}\big[\hat{\varepsilon}(\vx+\vdelta)^2\big]
= \big(\sum_{m=1}^M\alpha_m\big)^2\cdot \mathbb{E}_{h\sim\mathcal{H}}\big[\hat{\varepsilon}(\vx+\vdelta)^2\big]
\end{eqnarray*}
Since $\alpha_m\geq 0$ and $\sum_{m=1}^M\alpha_m=1$, thus, by definition,
\begin{eqnarray*}
\mathbb{E}_{h\sim\mathcal{H}}\big[(\bar{f}(\vx+\vdelta)-f(\vx+\vdelta))^2\big]\leq  \mathbb{E}_{h\sim\mathcal{H}}\big[(\hat{f}(\vx+\vdelta)-f(\vx+\vdelta))^2\big].
\end{eqnarray*}
This demonstrates that in the presence of assumption on the hypothesis space $\mathcal{H}$, the proposed objective $\bar{f}$ yields more transferable adversarial examples than existing method $\hat{f}$ with a higher probability.

\bibliography{mybibfile}
\bibliographystyle{elsarticle-num-names}
\end{document}